\documentclass[runningheads]{llncs}
\usepackage{graphicx}
\usepackage{amsmath,amssymb} 
\usepackage{color}
\usepackage{caption}
\usepackage{subcaption}
\usepackage[width=122mm,left=12mm,paperwidth=146mm,height=193mm,top=12mm,paperheight=217mm]{geometry}
\usepackage{url}
\newcommand{\smtext}[1]{\scriptsize{\textnormal{#1}}}
\begin{document}
\pagestyle{headings}
\mainmatter
\title{Detecting People in Cubist Art} 

\titlerunning{Detecting People in Cubist Art}

\authorrunning{Shiry Ginosar \and Daniel Haas \and Timothy Brown \and Jitendra Malik}

\author{Shiry Ginosar \and Daniel Haas \and Timothy Brown \and Jitendra Malik}
\institute{University of California Berkeley}

\maketitle

\begin{abstract}
Although the human visual system is surprisingly robust to extreme distortion when recognizing objects, most evaluations of computer object detection methods focus only on robustness to natural form deformations such as people's pose changes. To determine whether algorithms truly mirror the flexibility of human vision, they must be compared against human vision at its limits. For example, in Cubist abstract art, painted objects are distorted by object fragmentation and part-reorganization, to the point that human vision often fails to recognize them. In this paper, we evaluate existing object detection methods on these abstract renditions of objects, comparing human annotators to four state-of-the-art object detectors on a corpus of Picasso paintings. Our results demonstrate that while human perception significantly outperforms current methods, human perception and part-based models exhibit a similarly graceful degradation in object detection performance as the objects become increasingly abstract and fragmented, corroborating the theory of part-based object representation in the brain.
\end{abstract}

\section{Introduction}
The human visual system is amazingly robust to abstractness of object representation. While we can recognize people in natural, realistic images such as camera snapshots, we are also easily able to identify human figures rendered in numerous forms of artistic depiction such as oil paintings, cartoons and line drawings, despite the fact that they frequently have little in common with a real human being in terms of texture, color and form. At the extreme, abstract artists intentionally push the envelope of human vision to the point at which objects are completely unrecognizable, yet we still find such distorted shapes reminiscent of the human form. Computer vision detection algorithms can also recognize objects outside the realm of natural images, though their models of the visual world may not always align with the human one~\cite{superhuman}. Since algorithmic object detection in computer vision is usually compared against humans using natural images, such differences are seldom apparent. However, if we seek to design algorithms that achieve the power and flexibility of the human visual system by mimicking it, we should attempt to align the visual models of the detectors we train with the human model. Moreover, we must also evaluate their correspondence on images that stretch the limits of human vision.

In this paper, we employ the abstract depiction of objects in the Cubist paintings of Picasso as one such test corpus. Cubist paintings depict radically fragmented objects as if they were seen from many viewpoints at once, breaking them into medium sized cube-like parts that appear out of their natural ordering and do not conform to the rules of perspective~\cite{1949}. Because the abstract objects present in Cubist art are not normally seen in nature, the human visual system must strain to recognize them. Since humans are usually still able to identify the depicted object, Cubism shows us that human perception does not rely on exact geometry and is tolerant to a rearrangement of mid-level object parts. However, findings from neuroscience show that this ability degrades as images become more scrambled or abstract~\cite{pmid20224810}\cite{Ishai2007319}. We use the fragmentation and reordering of parts in Cubist paintings as an example of extreme conditions for the human visual system. Our claim is that if a method mimics human perceptual performance well, then it should behave similarly in these (as well as other) extreme conditions. We therefore aim to test whether there are existing detection methods that behave like human vision under these conditions. This stands in stark contrast to the common practice of evaluating computer vision models on datasets consisting only of camera snapshots and represents a novel contribution, as there is little research into how current systems perform on novel input. 

We choose to focus on part-based detection methods that permit the rearrangement of medium-complexity parts, as they have been proven to do well at representing naturally occurring form deformations~\cite{Felzenszwalb:2010ez}, and evaluate them in comparison to both human participants and object-level detection methods. Moreover, in order to chart the performance of the methods as human vision approaches its limit, we ask participants to divide the paintings into subsets according to the level of their abstractness and compare the performance of the humans and detection methods on each subset. Our results show that (1) existing part-based methods are relatively successful at detecting people even in abstract images, (2) that there is a natural correspondence between user ratings of image abstraction in the Cubist sense and part-based method performance, and (3) that these properties are not nearly as evident in non-part-based methods. By demonstrating that part-based methods mimic human performance, we both show that these methods are valuable for object recognition in non-traditional settings, and corroborate the theory of part-based object representation in the brain.

\section{Related Work}

Since in most tasks the human visual system serves as an upper bound benchmark for computer vision, some studies focus on characterizing its capabilities at the limit. For instance, Sinha and Torralba examined face detection capabilities in low resolution, contrast negated and inverted images~\cite{Sinha:2002un}. In other cases, computational models are used to test the validity of theories from neuroscience~\cite{Ullman:2002ch}. We take inspiration from these studies and evaluate human object detection in man-made art in order to provide a less restrictive benchmark of robustness to form abstraction and deformation than natural images. By using this benchmark we hope to discover parallels between the characteristics of human and algorithmic object detection.

From research in neuroscience, we know that the human visual system can detect and recognize objects even when they are manipulated in various ways~\cite{Lewis:2003wb}. For instance, humans are able to recognize inverted objects, although their performance is degraded, especially when the objects are faces or words~\cite{Tsao:2008cg}\cite{Sinha:2002un}. Similar results were obtained when comparing scrambled images to non-scrambled ones, leading to a theory of object-fragments rather than whole-object representations in the brain~\cite{GrillSpector:1998vg}\cite{Vogels:1999wh}. This theory is strengthened by recordings from neurons in the macaque middle face patch that indicate both part-based and holistic face detection strategies~\cite{Freiwald:2009kk}. Thus, although humans are capable of recognizing images distorted by scrambling, they are less adept at doing so. By analogy, we might expect methods trained on natural images to suffer a similar degradation in the face of a reorganization of object parts.

Object detection is one of the prominent unsolved problems in computer vision. Traditionally, object detection methods were holistic and template-based~\cite{DT05}, but recent successful detection methods such as Poselets~\cite{Bourdev:2009bl}\cite{Bourdev:2010jm} and deformable part-based models~\cite{Felzenszwalb:2010ez} have focused on identifying mid-level parts in an appropriate configuration to indicate the presence of objects. Other part-based methods discover mid-level discriminative patches in an unsupervised way~\cite{Singh2012DiscPat}\cite{doersch2012what}, use visual features of intermediate complexity for classification~\cite{Ullman:2002ch}, or rely on distinctive sparse fragments for recognition~\cite{AkselrodBallin:2008wc}. Finally, a model inspired by Cubism itself that assembles intermediate-level parts even more loosely has shown success in detecting objects~\cite{Nelson:1998tq}. Another approach to detection that has recently shown remarkable detection results is based on convolutional neural networks~\cite{girshick2014rcnn}\cite{sermanet-iclr-14}. We discuss the methods that we have chosen to benchmark in Section~\ref{sec:algorithms}.

\section{Cubist `fragments of perception'}

While there are many kinds of visual deformation to which human vision is robust, we choose to focus on the object fragmentation and part-reorganization exhibited by Cubist paintings as it has an appealing correlation with the strengths of part-based detection methods. Cubism is an art movement that peaked in the early 20th century with the work of artists such as Picasso and Braque. Cubist painters moved away from the two-dimensional representation of perspective characteristic of realism~\cite{1949}. Instead, they strove to capture the perceptual experience of viewing a three dimensional object from a close distance where in order to perceive an object, the viewer is forced to observe it part by part from different directions. Cubist painters collapsed these `fragments of perception' onto one two-dimensional plane, distorting perspective so that the whole object can be viewed simultaneously. Despite the abstraction of form, the original object is often readily detectable by the viewer as the parts chosen to represent it are naturalistic enough and discriminative enough to allow for the recognition of the object as a whole. However, this becomes harder with the degree of departure from reality~\cite{Ishai2007319}.

\begin{figure}
\vspace{-0.2in}
        \centering
        \begin{subfigure}[t]{2cm}
                \centering
                \includegraphics[height=3cm]{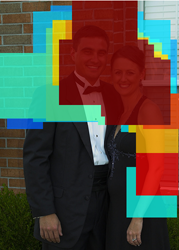}
        \end{subfigure}
        \quad
        \begin{subfigure}[t]{2cm}
                \centering
                \includegraphics[height=3cm]{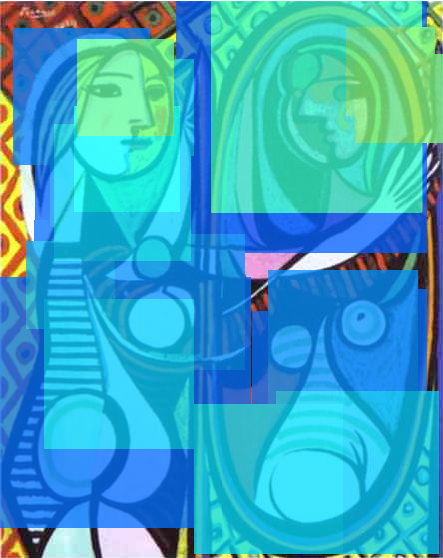}
        \end{subfigure}
        \caption{Heat maps showing the discriminative patches activations on a natural training image (Left) and ``Girl Before a Mirror 1932", a Picasso Cubist painting (Right). The color palette correlates with confidence score and ranges from blue (lowest) to red (highest). In both cases the most discriminative patches for class person are parts of faces and upper bodies, suggesting that computer vision methods are able to identify the key parts of human figures even when they are split into `fragments of perception' in Cubist paintings.}
        \label{fig:heatmap}
\vspace{-0.2in}
\end{figure}

\begin{figure}
\vspace{-0.1in}
        \centering
        \includegraphics[width=0.75\textwidth]{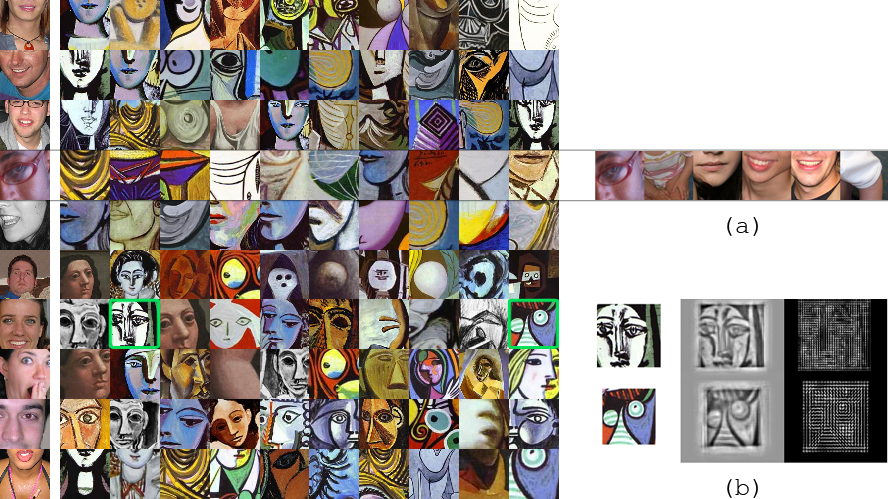}
        \caption{Discriminative patch detectors trained on natural images are able to detect the parts that characterize person figures in Cubist paintings. (Left) Each row displays the top ten discriminative patches activations\protect\footnotemark. The leftmost column shows the top activation on the training data. Most of the detectors find corresponding face parts in the natural and painting images, although many are false positives. The fourth patch-detector from the top detects patches with little visual consistency on the paintings as well as the training data (a). Some false positive activations (b)(Bottom) seem more similar to true positives (b)(Top) in Hoggles space~\cite{vondrick2013hoggles} (b)(Middle) than in HOG (b)(Right) or the original RGB (Left, marked in green) spaces.}
        \label{fig:all_patches}
\vspace{-0.1in}
\end{figure}

The fact that humans can detect non-figurative objects in Cubist paintings without prior training makes these paintings well-suited to benchmark robustness to abstractions of form in detection methods trained on natural images. In order to provide intuition that part-based models will be able to perform well on this task, we provide some initial evidence that computer vision methods can successfully identify key parts in the Cubist depictions of objects. We train an unsupervised discovery method of mid-level discriminative patches on the PASCAL 2010 ``person" class images~\cite{Singh2012DiscPat}\cite{doersch2012what}\cite{pascal-voc-2010}, and compare the part-detector activations on natural training images and Cubist paintings by Picasso in Figures~\ref{fig:heatmap} and~\ref{fig:all_patches}. Despite the difference in low-level image statistics, the detectors are able to discover the patches that discriminate people from non-people in both image domains. In the rest of the paper, we build on these results to test whether part-based object models can use the detected parts in order to recognize the depicted objects as a whole.

\footnotetext{For each detector, the 10 most confident activations are presented by decreasing confidence, excluding lower confidence duplicates of 50\% overlap or more. Detectors are sorted by the average score of their 20 top activations, excluding detectors where over $1/4$ of activations are duplicates of activations from higher rated detectors.}

\section{Object Detection Methods in Comparison}

\label{sec:algorithms}
We chose four person detectors that represent the range of available approaches to test whether state-of-the-art detection methods mimic the human visual system at its limits. These are presented in the time ordering in which they were proposed: one holistic template-based method, one part-based model where the parts are learned automatically, one part-based model where the parts are learned from human annotations, and the most recent deep learning method. Here we discuss the details of each one.

\textbf{Dalal and Triggs:} Object appearance in images can be characterized by histograms of orientations of local edge gradients binned over a dense image grid (HOG)~\cite{DT05}. The Dalal and Triggs (D\&T) method trains an object-level HOG template for detection using bounding box annotations. Since the features are binned, the detector is robust only to image deformation within the bins.

\textbf{Deformable Part Models:} A holistic HOG template cannot recognize objects in the face of non-rigid deformations such as varied pose, which result in a rearrangement of the limbs versus the torso. Therefore, the deformable part-based models detection method (DPM) represents objects as collections of parts that can be arranged with respect to the root part~\cite{Felzenszwalb:2010ez}. In practice, the model trained on natural images often learns sub-part models (such as half a face).

\textbf{Poselets:} Poselets is a similar HOG-based part model that considers extra human supervision during training~\cite{Bourdev:2009bl}\cite{Bourdev:2010jm}. Here, parts are not discovered but learned from body-part annotations. Poselets do not necessarily correspond to anatomical body parts like limbs or torsos as these are often not the most salient features for visual recognition. In fact, a highly discriminative Poselet for person detection corresponds to ``half of a frontal face and a left shoulder"~\cite{Bourdev:2009bl}.

\textbf{R-CNN:} R-CNN replaces the earlier rigid HOG features with features learned by a deep convolutional neural network~\cite{girshick2014rcnn}\cite{voc-release5}. While R-CNN does not have an explicit representation of parts, it is trained under a detection objective to be invariant to deformations of objects by using a large amount of data. Deep methods outperform previous algorithms by a large margin on natural data. Here we test this state-of-the-art method on abstract paintings.

\section{Experimental Setup}
\label{sec:setup}
Since we are interested in comparing the above methods to human vision on data that approaches the limit of perception, we conduct two kinds of experiments. First, we study the human and algorithm performance on person detection over our full corpus of Cubist paintings. Second, we examine the degradation in performance of human perception and detectors as the paintings become more abstract. We conduct all comparisons using the PASCAL VOC evaluation mechanism, in which true positives are selected based on a $50\%$ overlap between detection and ground truth bounding box~\cite{pascal-voc-2010}.

\subsection{Picasso Dataset}
In the experiments described below we used as our test data a set of 218 Picasso paintings that have titles indicating that they depict people. These ranged from figurative portraits to abstract depictions of person figures as a collection of distorted parts. The set of paintings we used is highly biased in comparison to PASCAL person class images~\cite{pascal-voc-2010}. Given the nature of the art form, Cubist paintings usually depict people in full frontal or portrait views where most of the canvas area is devoted to the torso of a person. This results in higher average precision scores as a random detection that contains over $50\%$ of the image would count as a true positive. This issue exists in any PASCAL VOC evaluation, but it is especially pronounced in this case.

\subsection{Human Perception Study Setup}
\label{sec:human_setup}
We conducted two experiments as part of our perception study. First, we recorded human detections of person figures in Cubist paintings. Second, we asked participants to bucket the paintings by their degree of abstraction compared to photorealistic depictions of people. For each painting, raters were asked to pick a classification on a 5-point Likert scale, where 1 corresponded to ``\textit{The figures in this painting are very lifelike}" and 5 corresponded to ``\textit{The figures in this painting are not at all lifelike}".

\vspace{-0.1in}
\subsubsection{Participants}
We recruited eighteen participants to partake in our perception study. Sixteen participants were undergraduate students at our institution, one was a graduate student and one a software engineering professional. Seventeen participants were male and one was female.

\vspace{-0.1in}
\subsubsection{Mechanism}
Participants completed the study on their personal laptops using an online graphical annotation tool we wrote for the purpose. Each participant spent an hour on the study and received a compensation of \$15. Each participant annotated 146 randomly chosen paintings out of the total 218, so that every painting was annotated by 14 - 15 unique participants.

\subsection{Detector Study Setup}
We compare the human recognition performance we measured during the perception study to four object detection methods. We train all methods using the PASCAL 2010 ``person" class training and validation images~\cite{pascal-voc-2010}. We train the methods using natural images so that they do not enjoy an advantage over humans by training on the paintings. However, some research suggests that human recognition in Cubist paintings does improve with repeated exposures~\cite{pmid20224810}. We set all parameters in all four methods to the same settings used in the original papers, except for the Poselets detection score which we set to 0.2 based on cross validation on the training data.

\subsection{Ground Truth}
Because Picasso did not explicitly label the human figures in his paintings, there is no clear cut gold standard for human figure annotations in our image corpus. As a result, we rely on our human participants to form a ground truth annotation set. We do so by capturing the average rater annotation as follows. Since each painting might have more than one human figure, we use k-means clustering to group annotations by the human figure they correspond to. For each cluster, we obtain a ground truth bounding box by taking the median of each corner of the bounding boxes in that cluster along every dimension. This yields one ground truth bounding box per human figure per image, which we can now use to evaluate both human and detector annotations. For each human rater we withhold her annotations and construct a modified leave-one-out ground truth from the annotations of all other raters. When evaluating detectors, however, we include annotations from all human raters in the ground truth.

It is worth noting that since humans themselves are error-prone (especially when recognizing objects in more abstract paintings), our ground truth cannot be a perfect oracle. Rather, all evaluation is comparing performance to the average, imperfect human. From this perspective, our evaluation of human raters can be seen as a measure of their agreement, and our evaluation of detectors can be seen as a measure of similarity to the average human.

\subsection{Evaluating Humans and Detectors}
Unlike detectors, humans provide only one annotation per figure without confidence scores, so we cannot compute average precision. Our primary metric for humans is the F-measure (F1 score), which is the harmonic mean of precision and recall. In order to combine the F-measures of all participants, we consider both (qualitatively) the shape of the distribution of the scores, and (quantitatively) the mean F-measure.

To compare detectors with humans, we pick the point on the methods' precision-recall curve that optimizes F-measure, and return the precision, recall, and F-measure computed at this point. This is generous to the detectors but captures their performance if they were well tuned for this task.

\section{Detection Performance on the Full Picasso Dataset}
\label{sec:performance}
In the first part of the comparison we evaluate the performance of the humans and the four methods at detecting human figures in Picasso paintings.

\subsection{Human Performance on the Full Picasso Dataset}
\begin{figure}
	\vspace{-0.1in}
	\centering
	\begin{subfigure}[t]{0.5\textwidth}
		\centering
		\includegraphics[height=2.5cm]{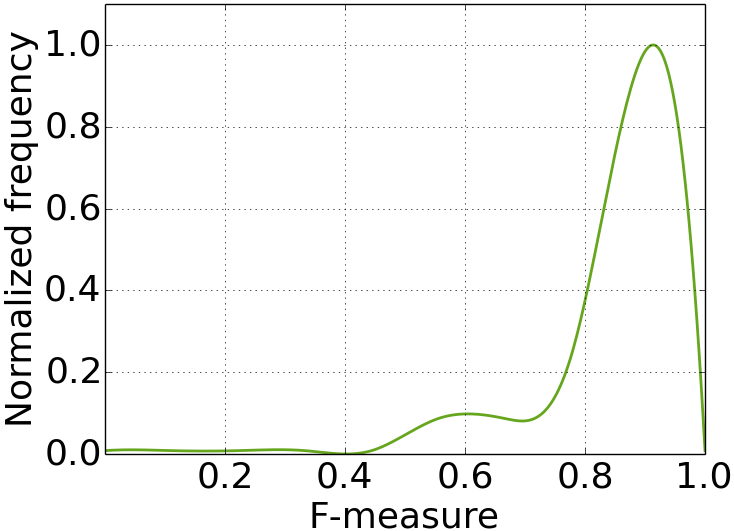}
	\end{subfigure}
	\caption{Frequency distribution of human F-measure recognition scores in all 218 Cubist paintings against the leave-one-out ground truth bounding boxes. Due to the small number of participants, the curve has been smoothed for clarity.}
	\label{fig:human_performance}
	\vspace{-0.35cm}
\end{figure}

First, we evaluate our human participants against the leave-one-out ground truth to determine how effective people are at recognizing human figures in Cubist art. Figure~\ref{fig:human_performance} displays the distribution of human F-measures for this task. Qualitatively, we see that humans perform quite well, as the distribution has its peak around 0.9, and there is little variance among the scores. It is worth noting the bump in the distribution around 0.6---there were a few raters whose annotations were significantly different from the ground truth annotation due to either a failure to recognize the images, or a misunderstanding of the annotation interface and instructions. Quantitatively, the first row of the table in Figure~\ref{fig:performance_results}(Right) shows the mean human precision, recall, and F-measure. These numbers confirm our impressions of the distribution---humans tend to agree with each other on the location of person figures in the paintings.

\subsection{Detector Performance on the Full Picasso Dataset}

\subsubsection{Qualitative Results}
\begin{figure}
	\vspace{-0.1in}
	\centering
	\begin{subfigure}[t]{0.25\textwidth}
		\centering
		\includegraphics[height=3cm]{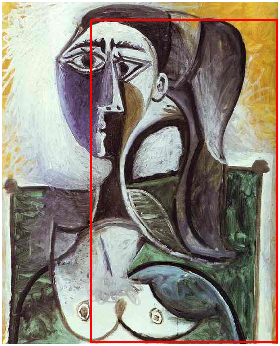}
	\end{subfigure}
	\begin{subfigure}[t]{0.4\textwidth}
		\centering
		\includegraphics[height=3cm]{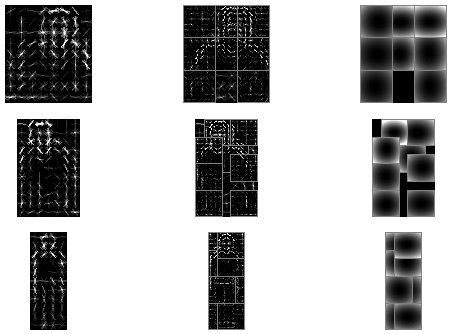}
	\end{subfigure}
	\caption{(Left) DPM detects one of the two viewpoints of the split face, resulting in a shifted localization of the person as a whole. (Right) The DPM model trained on natural images learns sub-face parts in three different scales. This provides insight as to why DPM is able to detect a split-face patch resulting in the bounding box detection on the left.}
	\label{fig:example_detections}
	\vspace{-0.3in}
\end{figure}

\begin{figure}
	\vspace{-0.1in}
\centering
	\includegraphics{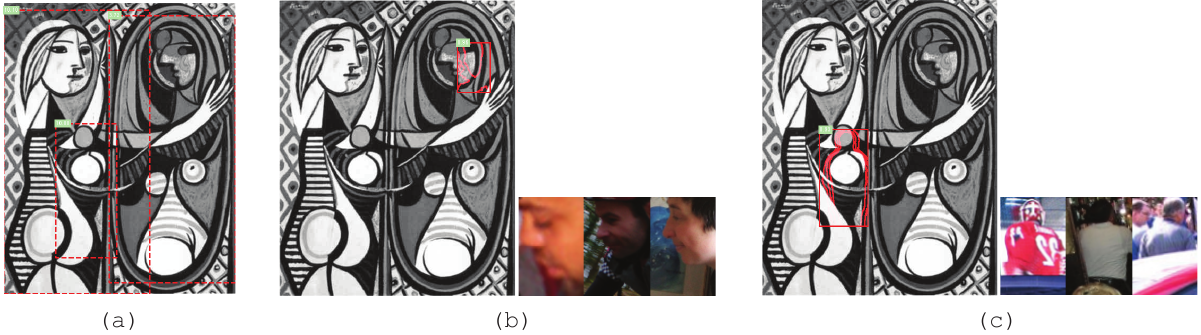}
\caption{The Poselets method is able to find person-parts in Cubist paintings and use them to detect person figures as a whole. (a) Bounding box detections in Picasso's ``Girl Before a Mirror 1932" with a false positive detection in the center. (b) A true positive Poselet activation on the painting (Left) together with the corresponding activation on the training data (Right). In both image domains the Poselet detects a downward-angled face in profile. (c)(Left) The false positive activation that results in the incorrect bounding box detection in (a). Here the Poselet falsely detects an away-facing person in the painting (c)(Right).} 
\vspace{-0.3in}
\label{fig:poselet_activation}

\end{figure}

\begin{figure}
	\vspace{-0.1in}
\centering
\includegraphics[width=\textwidth]{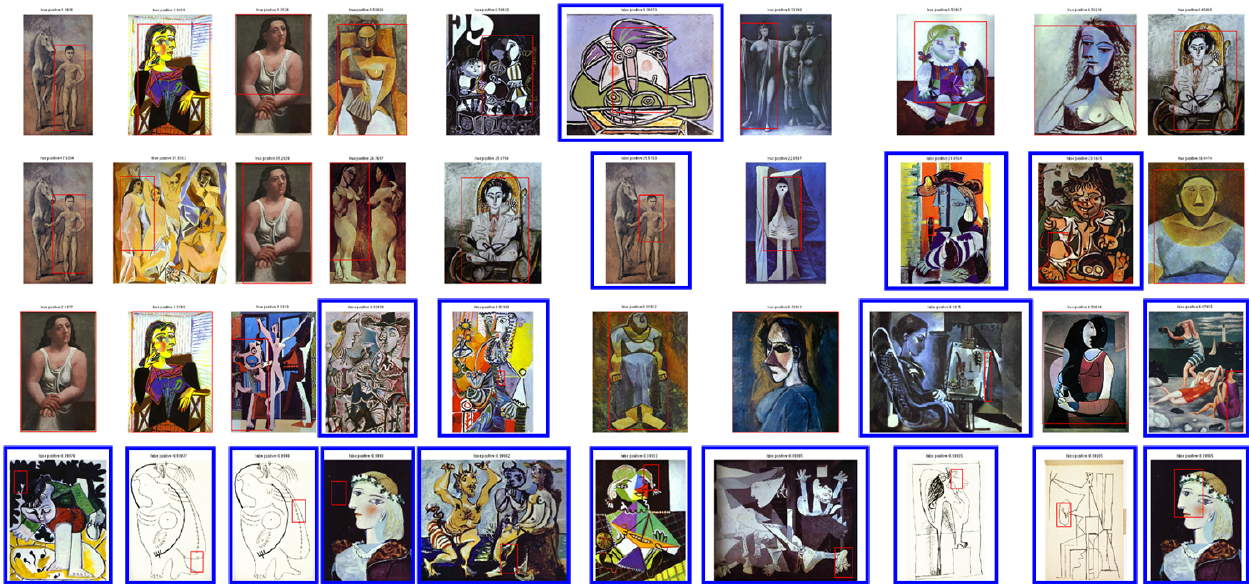}
\caption{Top ten detections for each method according to confidence from left to right. First row: DPM. Second row: Poselets. Third row: R-CNN. Fourth row: D\&T. False positives are marked in blue. Qualitatively, it is evident that part-based models outperform the other approaches.}
\label{fig:top10}
\vspace{-0.1in}
\end{figure}

Part-based models trained purely on photographs performed surprisingly well when tested against the Picasso data. As can be seen in Figures~\ref{fig:example_detections} and Figure~\ref{fig:poselet_activation}, Poselets and DPMs successfully produce bounding boxes for figures in the paintings, though they have their fair share of false positives and misses. The non-part-based methods do not perform as well, as is evident in Figure~\ref{fig:top10} where each row displays the top ten detections of a method over the entire painting dataset, sorted by each method's confidence score. 

\subsubsection{Quantitative Results}
\begin{figure}
	\vspace{-0.1in}
  \begin{minipage}[b]{0.5\textwidth}
    \centering
   \begin{subfigure}[b]{\textwidth}
		\centering
		\includegraphics[height=4cm]{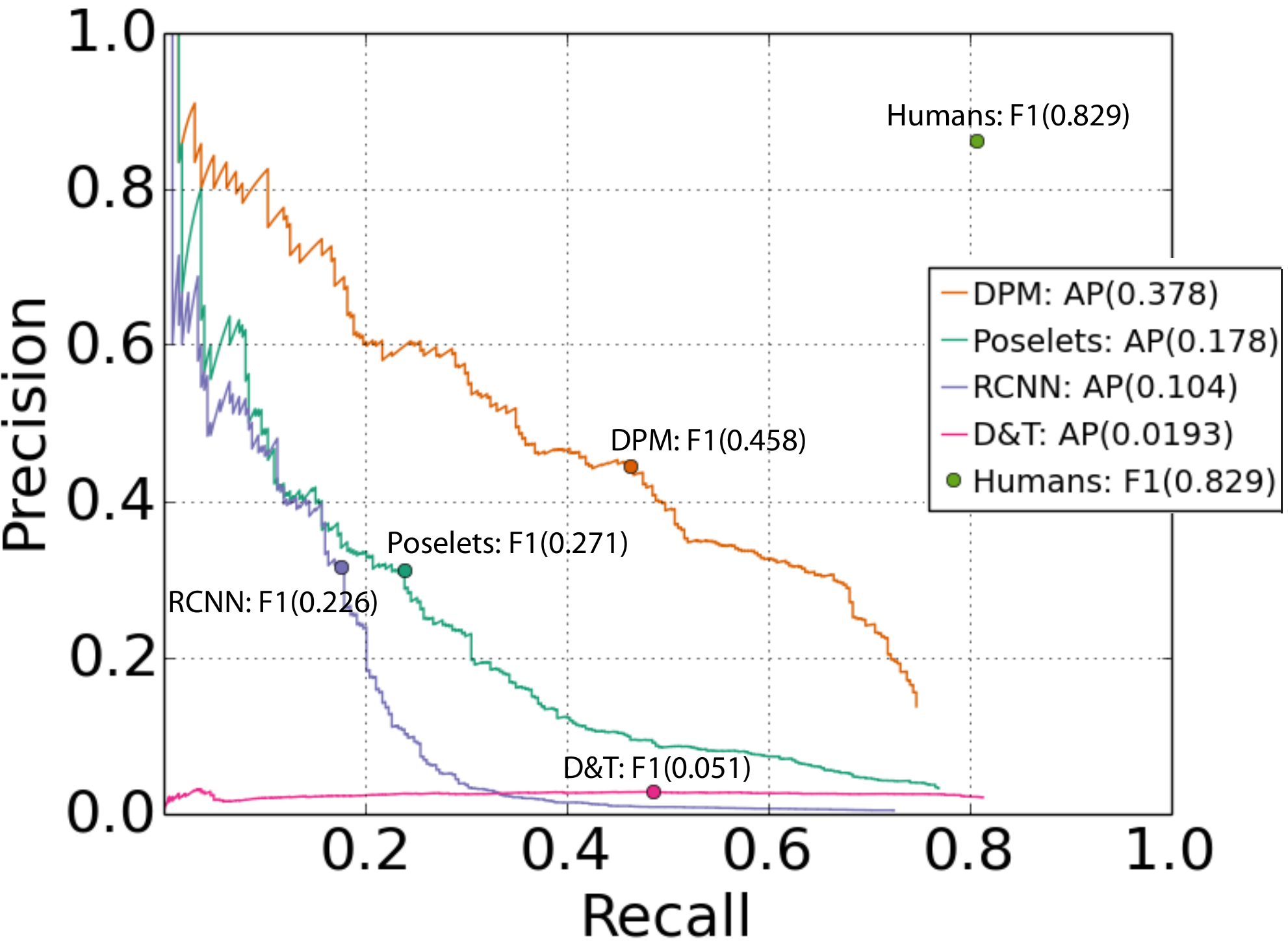}
	\end{subfigure}
     \end{minipage}
  \begin{minipage}[b]{0.5\textwidth}
    \centering
\begin{tabular}[b]{c|c|c|c|c}
\textbf{\smtext{Annotator}} & \textbf{\smtext{Precision}} & \textbf{\smtext{Recall}} & \textbf{\smtext{F-measure}} & \textbf{\smtext{AP}} \\
\hline
\smtext{Human} & \smtext{0.804} & \smtext{0.860} & \smtext{0.829} & \smtext{N/A} \\
\smtext{DPM} & \smtext{0.444} & \smtext{0.464} & \smtext{0.458} & \smtext{0.378}\\
\smtext{Poselets} & \smtext{0.311} & \smtext{0.240} & \smtext{0.271} & \smtext{0.178}\\
\smtext{RCNN} & \smtext{0.315} & \smtext{0.177} & \smtext{0.226} & \smtext{0.104}\\
\smtext{D\&T} & \smtext{0.027} & \smtext{0.486} & \smtext{0.051} &  \smtext{0.019} 
\end{tabular}
\end{minipage}
\caption{Performance comparison via precision-recall curves (Left) and tabular data (Right). In both the plot and the table: for detectors, precision, recall, and F-measure are the maxima over the entire precision-recall curve. For humans, these numbers are averages across the raters. While DPM outperforms other methods, none of the methods reach human performance.}
\label{fig:performance_results}
\vspace{-0.2in}
\end{figure}

Figure~\ref{fig:performance_results}(Left) displays the precision-recall curves for each method when evaluated on all paintings. There is a clear ordering in accuracy: DPM performs the best by far, Poselets and RCNN come next, and Dalal and Triggs, though characterized by high recall, has extremely low precision, leading to terrible performance. In general, all of the methods achieve recall of up to 0.8, which implies that they are capable of recognizing most human figures in the image. However, DPM is the only method which can maintain a reasonable precision as recall increases, which explains its significantly greater AP. The table in Figure~\ref{fig:performance_results}(Right) confirms these insights. DPM's performance on this task is quite encouraging, as its AP (0.38) is not too far off from its performance on photorealistic PASCAL 2010 photographs (0.41 without context rescoring)~\cite{voc-release5}. In practice this a favorable comparison as the PASCAL dataset, unlike ours, includes many images that do not contain people.
\vspace{-0.1in}

\subsection{Comparing Human and Detector Performance}

As is clear from the graphical and tabular data in Figure~\ref{fig:performance_results}, a pair of humans are much better than a human and a detector at reaching an agreement about where the person figures are in Picasso paintings. The green dot in the upper right corner of the graph shows human precision and recall to be far higher than the orange dot of DPM, the highest-performing method.

\subsection{Discussion of Performance on the Full Picasso Dataset}

The comparison between the human participants and four methods on the task of detecting person figures in Picasso paintings demonstrates clearly that humans are highly skilled at this task, and that detectors are much less effective but can still achieve results within an order of magnitude of their detection performance on natural images. Among algorithms, part-based object detection methods  perform better than object-level methods on images containing form abstraction. DPM and Poselets, our two part-based methods, demonstrate the best performance on the object detection task. This is likely due to the fact that part-based methods are able to recognize medium-level parts that remain intact even in Cubist paintings where standard human body parts are highly fragmented and rearranged. We emphasize the part-based approach here, as we have no reason to believe that the HOG features used by both DPM and Poselets carry any advantage over other image features organized in a part-based model.

Given its success in object detection on natural images, it is interesting that R-CNN does not perform well on this task. One reason for this could be that R-CNN is not a part-based model, however this is only partly true because the convolutional filters can be thought of as parts, and the max pooling as performing deformations. A second factor might be the fact that R-CNN over fits to the natural visual world and fails at adapting to the domain of paintings. There has been little research into how CNN-based networks perform on distorted images, but an initial investigation suggests that tiny changes to an image may cause drastic changes to their output~\cite{DBLP:journals/corr/SzegedyZSBEGF13}.

\section{Performance Degradation with Increased Abstraction}
\label{sec:degradation}
In the second part of the comparison we study the degradation of performance of humans and methods with increased painting abstractness.
	\vspace{-0.1in}

\subsection{Classifying Images by Degree of Abstraction}
In our user study (described in Section~\ref{sec:human_setup}), each rater labeled images on a scale from 1 (\textit{The figures in this painting are very lifelike}) to 5 (\textit{The figures in this painting are not at all lifelike}). By taking the rounded average of user labels for each image, we divide the images into five `buckets of abstraction' in order to evaluate object detection performance as the abstraction of form increases. An example of a painting with an average rating of 1 is Picasso's ``Seated Woman 1921", and an example of a painting with an average rating of 5 is Picasso's ``Nude and Still Life 1931" (copyrighted paintings not reproduced). The number of images in each bucket is shown in Figure~\ref{fig:degradation_results}(Top Left).

\subsection{Human Performance Degradation}
\begin{figure}
\vspace{-0.3in}
	\centering
	\includegraphics[height=8cm]{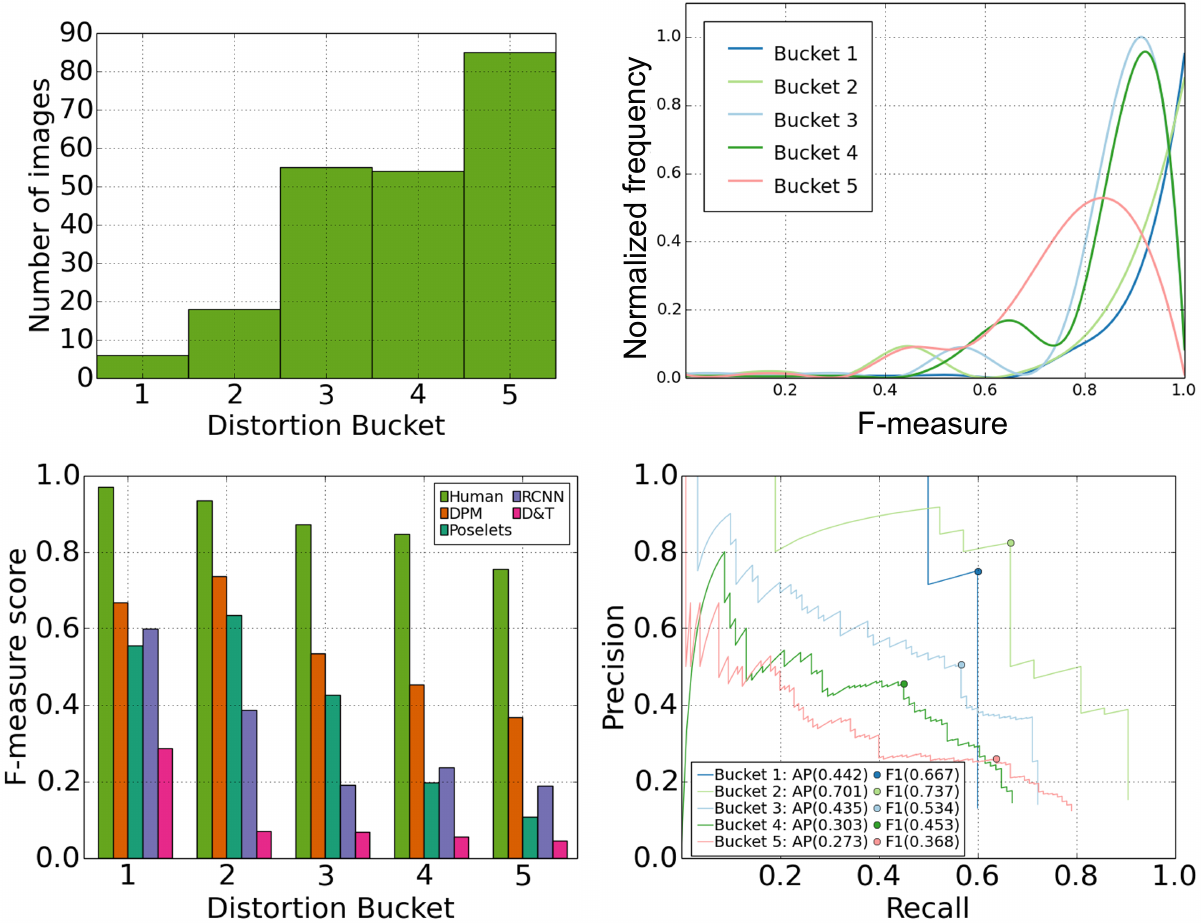}
	\caption{Impact of increasing form abstraction on object detection performance. Bucket 1 contains images that appear most natural, and bucket 5 contains images that appear most abstract. (Top Left) A histogram showing the number of images per bucket. (Top Right) Human F-measure distributions by abstraction bucket. (Bottom Right) DPM precision-recall curves by bucket. (Bottom Left) Comparing humans and methods. Part-based models show a similar degradation behavior to human performance as the images become more abstract.}
	\label{fig:degradation_results}
	\vspace{-0.2in}
\end{figure}

Figure~\ref{fig:degradation_results}(Top Right) demonstrates the impact of abstraction of form on human object detection performance. As the images become more abstract, the distribution of human F-measures shifts clearly to the left. This indicates that human performance worsens on more scrambled human figures, which is consistent with previous results\!~\cite{Tsao:2008cg}\cite{Sinha:2002un}. As noted in Section~\ref{sec:performance}, there are a few annotators with low F-measures in each of the curves, which implies that these raters' errors are independent of image abstraction and are most likely due to a failure to follow the instructions rather than an inability to recognize objects.

\subsection{Detector Performance Degradation}

\subsubsection{Qualitative Results}
\begin{figure}
	\vspace{-0.1in}
	\centering
	\includegraphics[width=\textwidth]{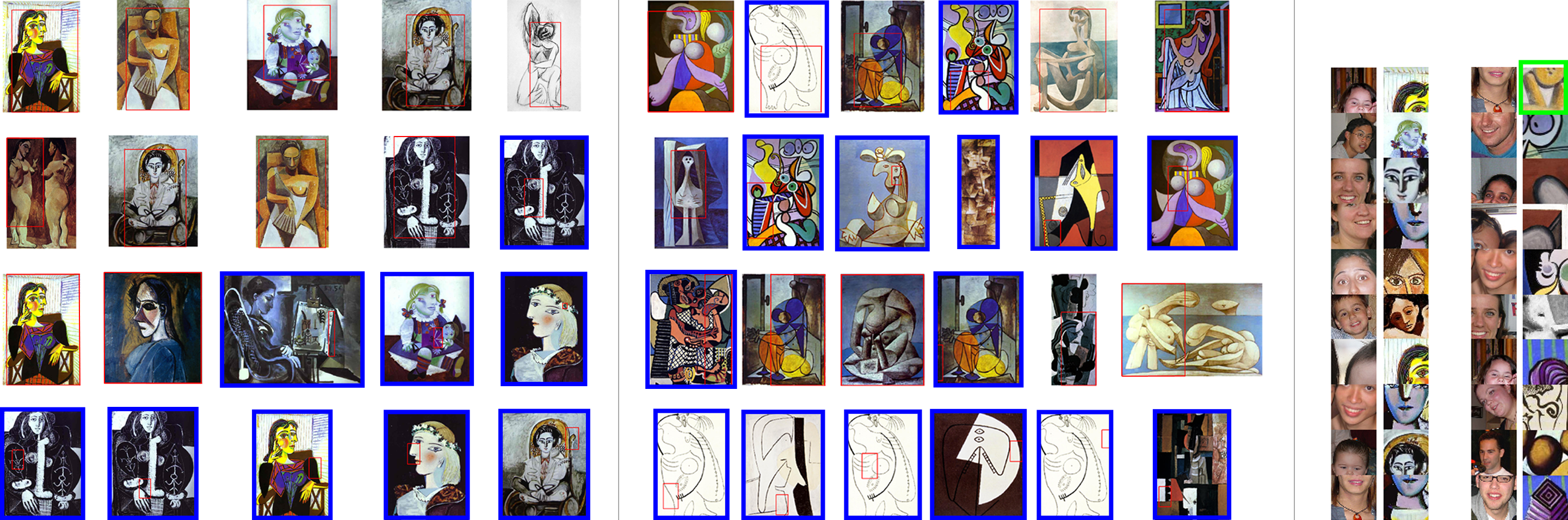}
	\caption{The degradation in performance with image difficulty. Top five detections per method (rows correspond to: DPM, Poselets, R-CNN, D\&T) on images from bucket 2 (Left) and bucket 5 (Middle). False positives are marked in blue. This comparison shows that all detection methods perform worse on more abstract images. (Right) Degradation in performance is also evident in the detection of isolated parts. Top 10 discriminative patches activations for images of bucket 2 and bucket 5, with corresponding activations on PASCAL images. All activations on bucket 2 are true positives compared to only one from bucket 5 (in green).  }
	\label{fig:detections_degradation_montage}
	\vspace{-0.2in}
\end{figure}

In Figure~\ref{fig:detections_degradation_montage} we compare the top detections per method on paintings from bucket 2 versus bucket 5. The top discriminative-patches activations on these two buckets (Right) help visualize the difficulty in detecting meaningful mid-level parts as the paintings become more abstract.

\vspace{-0.1in}
\subsubsection{Quantitative Results}
Figure~\ref{fig:degradation_results}(Bottom Right) shows the precision recall curves for the DPM method with varying abstraction buckets. As with human performance, increasing the image abstraction of form causes a pronounced decrease in performance. The overlap between the curves for buckets 1 and 2 may be due to variance as a consequence of the low number of images in those two buckets.
	\vspace{-0.1in}

\subsection{Comparing Human and Detector Degradation}
Figure~\ref{fig:degradation_results}(Bottom Left) compares the performance change of humans and detectors with varying abstraction buckets. As can be seen, the performance of all detectors degrades in a similar pattern to human performance as images become increasingly abstract, but the part-based methods follow the human pattern most closely. This matches our intuition about the similarities between part-based object detection and the human visual system. In contrast, the template-based Dalal and Triggs method abruptly breaks down after bucket 1.

\subsection{Discussion of Degradation with Increased Abstraction}
As we have demonstrated, part-based models for object detection show a smooth degradation in precision and recall as the images become more abstract. This is consistent with results from neuroscience, 
which indicate that humans are capable of detecting objects cut into parts, but that their ability degrades significantly when the parts are scrambled. The correspondence between human and computational method performance on this task suggests that a part-based object representation might be a good approximation for the mechanisms of object detection in the human brain. The ability to model these mechanisms computationally further corroborates the neuroscience theory of part-based object detection strategies. This is encouraging, even though current methods cannot yet perform at the level of human vision.

We note that the correspondence between part-based models and human perception is a somewhat surprising one. At their core, the methods we used are based on HOG features that we expected to be highly dependent on image statistics. It was pleasantly surprising to observe the correlation between these methods, trained on natural images, and human perception on Cubist paintings with completely different statistics. We believe that better performance could be achieved using part-based models that rely on higher level features than HOG.

\section{Conclusions}
Since computer vision aims to attain not only the performance of human vision but also its flexibility and robustness, we should characterize our algorithms' performance on novel and extreme inputs. In this paper, we have argued that object detection under abstraction of object form is an example of a challenging perception task that existing image benchmarks do not properly evaluate. We have proposed Cubist paintings as an additional corpus for object detection, as they contain rearranged object parts that are nevertheless recognizable to the human visual system as whole objects. Using this dataset, our evaluation comparing human performance to that of various object detection methods demonstrates that part-based models are a step in the right direction for modeling human robustness to part-rearrangement, since their performance degrades comparably to humans as the abstractness in images increases. By showing that these models can be trained on photographic data yet still perform on abstract data we demonstrate that they are less over-fit to the natural world than template-based and deep models.

Part-reorganization in Cubism is one example that pushes the envelope of human perception, but there are other artistic movements with characteristic abstractions, such as the use of blurring in Impressionism, that would provide rich grounds for study. Future work in the design of computer vision methods should be cognizant of the limitations of traditional camera snapshot datasets and look for complementary resources when evaluating computational methods. Ultimately, a plethora of such resources should not only be used for testing, but must be combined into rich training datasets when designing methods that truly mimic the wide range of human perception.

\section{Acknowledgments}
The authors would like to thank Mark Lescroart for his guidance and advice throughout this project,  Bharath Hariharan, Carl Doersch, Katerina Fragkiadaki and Emily Spratt for their insightful comments and Sean Arietta for his code. This material is based upon work supported by the National Science Foundation Graduate Research Fellowship under Grant No. DGE 1106400.

\bibliographystyle{splncs03}
\bibliography{egbib}

\begin{thebibliography}{10}
\providecommand{\url}[1]{\texttt{#1}}
\providecommand{\urlprefix}{URL }

\bibitem{AkselrodBallin:2008wc}
Akselrod-Ballin, A., Ullman, S.: {Distinctive and compact features}. Image and
  Vision Computing  26(9),  1269--1276 (September 2008)

\bibitem{Bourdev:2009bl}
Bourdev, L., Malik, J.: {Poselets: Body part detectors trained using 3D human
  pose annotations}. In: Proceedings of the IEEE International Conference on
  Computer Vision ({ICCV}). pp. 1365--1372 (2009)

\bibitem{Bourdev:2010jm}
Bourdev, L.D., Maji, S., Brox, T., Malik, J.: {Detecting People Using Mutually
  Consistent Poselet Activations.} In: Proceedings of the European Conference
  on Computer Vision ({ECCV}). pp. 168--181 (2010)

\bibitem{DT05}
Dalal, N., Triggs, B.: Histograms of oriented gradients for human detection.
  In: Proceedings of the IEEE Conference on Computer Vision and Pattern
  Recognition ({CVPR}). vol.~2, pp. 886--893 (2005)

\bibitem{doersch2012what}
Doersch, C., Singh, S., Gupta, A., Sivic, J., Efros, A.A.: What makes paris
  look like paris? ACM Transactions on Graphics (SIGGRAPH)  31(4) (2012)

\bibitem{pascal-voc-2010}
Everingham, M., Van~Gool, L., Williams, C.K.I., Winn, J., Zisserman, A.: The
  {PASCAL} {V}isual {O}bject {C}lasses {C}hallenge 2010 {(VOC2010)} {R}esults.
  http://www.pascal-network.org/challenges/VOC/voc2010/workshop/index.html

\bibitem{Felzenszwalb:2010ez}
Felzenszwalb, P., Girshick, R., McAllester, D., Ramanan, D.: Object detection
  with discriminatively trained part based models. Pattern Analysis and Machine
  Intelligence ({PAMI})  32(9) (2010)

\bibitem{Freiwald:2009kk}
Freiwald, W.A., Tsao, D.Y., Livingstone, M.S.: {A Face Feature Space in the
  Macaque Temporal Lobe.} Nature Neuroscience  12(9),  1187--1196 (Sep 2009)

\bibitem{girshick2014rcnn}
Girshick, R., Donahue, J., Darrell, T., Malik, J.: Rich feature hierarchies for
  accurate object detection and semantic segmentation. In: Proceedings of the
  IEEE Conference on Computer Vision and Pattern Recognition ({CVPR}) (2014)

\bibitem{voc-release5}
Girshick, R.B., Felzenszwalb, P.F., McAllester, D.: Discriminatively trained
  deformable part models, release 5.
  \url{http://people.cs.uchicago.edu/~rbg/latent-release5/}

\bibitem{GrillSpector:1998vg}
Grill-Spector, K., Kushnir, T., Hendler, T.: {A sequence of object-processing
  stages revealed by fMRI in the human occipital lobe}. Human Brain Mapping
  6(4),  316--328 (1998)

\bibitem{superhuman}
Hsiao, E., Efros, A.A.: {DPM} superhuman.
  \url{http://www.cs.cmu.edu/~efros/courses/LBMV09/presentations/latent_presentation.pdf}
  slides 43-51

\bibitem{Ishai2007319}
Ishai, A., Fairhall, S.L., Pepperell, R.: Perception, memory and aesthetics of
  indeterminate art. Brain Research Bulletin  73(4–6),  319 -- 324 (2007)

\bibitem{1949}
Laporte, P.M.: Cubism and science. The Journal of Aesthetics and Art Criticism
  7(3),  pp. 243--256 (1949)

\bibitem{Lewis:2003wb}
Lewis, M.B., Edmonds, A.J.: {Face detection: Mapping human performance}.
  Perception  32(8),  903--920 (2003)

\bibitem{Nelson:1998tq}
Nelson, R.C., Selinger, A.: {A Cubist Approach to Object Recognition.} In:
  Proceedings of the IEEE International Conference on Computer Vision ({ICCV}).
  pp. 614--621 (1998)

\bibitem{sermanet-iclr-14}
Sermanet, P., Eigen, D., Zhang, X., Mathieu, M., Fergus, R., LeCun, Y.:
  Overfeat: Integrated recognition, localization and detection using
  convolutional networks. In: International Conference on Learning
  Representations (ICLR 2014). CBLS (2014)

\bibitem{Singh2012DiscPat}
Singh, S., Gupta, A., Efros, A.A.: Unsupervised discovery of mid-level
  discriminative patches. In: Proceedings of the European Conference on
  Computer Vision ({ECCV}) (2012)

\bibitem{Sinha:2002un}
Sinha, P., Torralba, A.: {Detecting faces in impoverished images}. Journal of
  Vision  2(7) (November 2002)

\bibitem{DBLP:journals/corr/SzegedyZSBEGF13}
Szegedy, C., Zaremba, W., Sutskever, I., Bruna, J., Erhan, D., Goodfellow,
  I.J., Fergus, R.: Intriguing properties of neural networks. CoRR
  abs/1312.6199 (2013)

\bibitem{Tsao:2008cg}
Tsao, D.Y., Livingstone, M.S.: {Mechanisms of face perception.} Annual Review
  of Neuroscience  31,  411--437 (2008)

\bibitem{Ullman:2002ch}
Ullman, S., Vidal-Naquet, M., Sali, E.: {Visual features of intermediate
  complexity and their use in classification.} Nature Neuroscience  5(7),
  682--687 (Jul 2002)

\bibitem{Vogels:1999wh}
Vogels, R.: {Effect of image scrambling on inferior temporal cortical
  responses.} Neuroreport  10(9),  1811--1816 (Jun 1999)

\bibitem{vondrick2013hoggles}
Vondrick, C., Khosla, A., Malisiewicz, T., Torralba, A.: {HOGgles: Visualizing
  Object Detection Features}. In: Proceedings of the IEEE International
  Conference on Computer Vision ({ICCV}) (2013)

\bibitem{pmid20224810}
Wiesmann, M., Ishai, A.: {{T}raining facilitates object recognition in cubist
  paintings}. Frontiers in Human Neuroscience  4, ~11 (2010)

\end{thebibliography}
\end{document}